\documentclass[conference]{IEEEtran}

\usepackage{booktabs}
\usepackage{graphicx}
\usepackage{tabularx}
\usepackage{amsmath}
\usepackage{amssymb}
\usepackage{textcomp}
\usepackage{enumitem}
\usepackage{booktabs}
\usepackage{multirow}
\usepackage{siunitx}
\usepackage{balance}
\usepackage{xcolor}
\usepackage{cite}
\usepackage[para,online,flushleft]{threeparttable}
\usepackage[linesnumbered, noend, algoruled,boxed,lined]{algorithm2e}

\ifCLASSINFOpdf
\else
\fi

\begin{document}
\title{Towards Efficient and Scalable Acceleration of Online Decision Tree Learning on FPGA}

\author{\IEEEauthorblockN{Zhe Lin\IEEEauthorrefmark{1}, Sharad Sinha\IEEEauthorrefmark{2} and Wei Zhang\IEEEauthorrefmark{1}}
	\IEEEauthorblockA{\IEEEauthorrefmark{1}Hong Kong University of Science and Technology, Hong Kong}
	\IEEEauthorblockA{\IEEEauthorrefmark{2}Indian Institute of Technology (IIT) Goa, India\\
		zlinaf@ust.hk, sharad\_sinha@ieee.org, wei.zhang@ust.hk}}

\maketitle

\begin{abstract}
Decision trees are machine learning models commonly used in various application scenarios. In the era of big data, traditional decision tree induction algorithms are not suitable for learning large-scale datasets due to their stringent data storage requirement. Online decision tree learning algorithms have been devised to tackle this problem by concurrently training with incoming samples and providing inference results. However, even the most up-to-date online tree learning algorithms still suffer from either high memory usage or high computational intensity with dependency and long latency, making them challenging to implement in hardware. To overcome these difficulties, we introduce a new quantile-based algorithm to improve the induction of the Hoeffding tree, one of the state-of-the-art online learning models. The proposed algorithm is light-weight in terms of both memory and computational demand, while still maintaining high generalization ability. A series of optimization techniques dedicated to the proposed algorithm have been investigated from the hardware perspective, including coarse-grained and fine-grained parallelism, dynamic and memory-based resource sharing, pipelining with data forwarding. We further present a high-performance, hardware-efficient and scalable online decision tree learning system on a field-programmable gate array (FPGA) with system-level optimization techniques. Experimental results show that our proposed algorithm outperforms the state-of-the-art Hoeffding tree learning method, leading to 0.05\% to 12.3\% improvement in inference accuracy. Real implementation of the complete learning system on the FPGA demonstrates a 384$\times$ to 1581$\times$ speedup in execution time over the state-of-the-art design.
\end{abstract}
	
\IEEEpeerreviewmaketitle

\vspace{-4mm}
\section{Introduction}
Decision tree algorithms are a popular class of machine learning algorithm and have been deployed in many real scenarios~\cite{cvprchen16, kaneko18, oliv18}, especially when multiple decision trees are combined into powerful ensemble models, such as XGBoost~\cite{xgboost} and random forests~\cite{brei01}. Recently, the ensemble of decision trees as deep forests~\cite{zhou17} has been reported to produce comparable performance compared to deep neural networks. However, there are several drawbacks that limit the full exploitation of the traditional decision trees (e.g., IDT3~\cite{idt3}, CART~\cite{cart} and C4.5~\cite{quin14}). The first drawback is the extensive memory consumption during the training process, which is proportional to the size of datasets. Classic decision tree learners assume that the complete datasets can be preloaded before training starts. This reduces their capability to train with large-scale datasets, especially when, nowadays, large amount of data is being generated daily. The second disadvantage comes with the learners' inability to adapt themselves to new data once the training process is terminated. In the era of big data, the size of datasets is no longer the bottleneck of learning algorithms. Instead, the ability to effectively learn from massive data and rationally make use of incoming data becomes more fundamental and critical.

To broaden the applicability of decision tree algorithms, extensions from traditional tree algorithms to batch learning and online learning (or so-called incremental learning) have been studied, which aim at adapting the models to incoming data without losing previously learned knowledge. One of the state-of-the-art online learning methods for streaming data is the \textit{Hoeffding tree}~\cite{pedro00} algorithm and its variants~\cite{vfdtc, vfml, streamdm, ga08, hul01, ikno11, kour16, vasi17}. The Hoeffding tree presents an enhancement of the decision tree induction algorithm which leverages the accumulated samples to estimate complete datasets statistically. It is capable of performing training and inference concurrently. The Hoeffding tree is widely used in various application scenarios~\cite{bar14, fai15, wu16, nie17}.

While efficient software implementation has been investigated for processors to accelerate the Hoeffding tree~\cite{vfml, streamdm}, there are still many hindrances to the compact implementation and optimization of the Hoeffding tree design from the hardware perspective. We identify two principal challenges limiting Hoeffding tree implementation in hardware: 1) the high cost of memory usage to store the required subset of samples as well as characteristics in each leaf node; and 2) the high computational demand with dependency and long latency between iterations in the learning process, which can hamper efficient data processing with optimization schemes such as parallelism and pipelining. Furthermore, we observe a trade-off between the above two factors in the state-of-the-art designs: the methods in~\cite{streamdm} and~\cite{ga08}, attempting to reduce the memory usage, tend to extensively increase the computational intensity and latency, and vice versa, as in the proposed methods of~\cite{vfdtc} and~\cite{vfml}. The high and unbalanced need of memory and computation makes the existing approaches difficult to efficiently implement in hardware, especially on FPGAs where memory and digital signal processing (DSP) resources are both limited. Motivated by the above challenges and observations, we seek opportunities to implement and optimize the Hoeffding tree in a hardware-friendly and scalable way, and also strive to make use of resources in a more balanced manner. In this paper, we propose the first and complete implementation of the Hoeffding tree learning system on FPGA, with the following contributions:
\begin{itemize}
	\item We first introduce a quantile-based algorithm for Hoeffding tree induction, which uses light-weight computation and constant memory, while preserving high accuracy.
	\item We present hardware optimization techniques dedicated to the proposed algorithm, in order to achieve high hardware efficiency and scalability. These includes different levels of parallelism, dynamic and memory-based resource sharing, and pipelining with data forwarding.
	\item We investigate optimization techniques for tree growing, categorical attribute learning and split judgment to establish the complete online decision tree system on FPGA.
\end{itemize}

\section{Algorithm and Challenges}
\subsection{Hoeffding Tree Induction Algorithm}
The basic induction flow of the Hoeffding tree~\cite{pedro00} is the same as the conventional decision trees~\cite{rok08}, except that Hoeffding tree exploits the potential for the currently seen sample set to represent an infinite sample set. The Hoeffding tree algorithm is described in Algorithm~\ref{alg: ht}. \textit{Hoeffding bound} (additive Chernoff bound)~\cite{hoeffb} tells how close the current best split approaches the optimal split given an infinite sample set. Suppose we make $n$ independent observations of a random variable $ r$ within range $R$. The Hoeffding bound guarantees that the true mean $\overline{r}$ of $r$ will be at least $E[r]-\epsilon$, with
\begin{equation}
	\small
	\label{eq: hb}
	\epsilon = \sqrt{\frac{R^2ln(1/\delta)}{2n}}.
\end{equation}

Let $G(a_i)$ be the best measurement (e.g., gini impurity reduction) of a chosen split attribute $a_i$. The Hoeffding tree searches for the best and second-best $G(\cdot)$ values amongst all attributes. Given the sample set of size $n$ for a specific node and a desired $\delta$, the Hoeffding bound justifies that the current best attribute is the exact best attribute from an infinite dataset with probability $1-\delta$, if the following equation is satisfied:
\begin{equation}
	\small
	\label{eq: htb}
	G(Best\ attr.)-G(2^{nd}\ Best\ attr.)>\sqrt{\frac{R^2ln(1/\delta)}{2n}}.
\end{equation}

An additional tie condition is applied: when the two best attributes have close $G(\cdot)$, a split is taken if the Hoeffding bound is lower than a certain threshold $\tau$. That is,
\begin{equation}
	\small
	\label{eq: htbtie}
	G(Best\ attr.)-G(2^{nd}\ Best\ attr.) < \sqrt{\frac{R^2ln(1/\delta)}{2n}} < \tau.
\end{equation}

\subsection{Challenges}
\label{subsec: chalg}
Studies~\cite{vfdtc, vfml, streamdm, ga08} have introduced several methods to improve the basic Hoeffding tree algorithm. These methods, however, reveal two main challenges for hardware implementation.

\textbf{1. High Cost of Memory Utilization.} In VFML~\cite{vfml}, both numeric and categorical attribute values are preserved in a fixed number of bins (denoted as $n_{ijk}$) in a first-come-first-served manner. If all the bins are occupied, the newly coming attribute values unseen in all the bins are simply discarded during runtime. Although this method works well with categorical attributes of which values are discrete and the total number can be determined in the compile time, it requires a bin of large size to fit each numeric attribute per class per node to achieve a wide value coverage. Hence, the memory requirement grows significantly with the number of attributes. This similarly exists in the method~\cite{ga08} using Greenwald and Khanna summaries~\cite{green01}, which requires to construct sample distribution from up to thousands of tuples per attribute-class combination per node. The exhaustive binary tree method~\cite{vfdtc} also suffers from injudicious use of memory because it needs to dynamically allocate memory for sample storage. 

\LinesNumbered
\begin{algorithm}[t]
	\footnotesize
	\caption{Traditional Hoeffding tree algorithm}
	\label{alg: ht}
	\SetKwInOut{Input}{input}
	\SetKwInOut{Output}{output}
	\Input{samples denoted as $(x,y)$}
	\Output{Hoeffding tree denoted as $HT$}
	\For{each $(x_t, y_t)$ coming at time $t$}{
		filter $(x_t,y_t)$ to leaf $l$ of $HT$\\
		sample number in leaf $l$: $n_l \leftarrow n_l+1$\\
		update bin count $(attr_i, val_j, class_k)$ $n_{ijk}$ in leaf $l$\\
		\If{split trial is activated}{
			compute left/right partitions according to $n_{ijk}$\\
			compute $G(\cdot)$ for each attribute\\
			\If{$G(best)$\ -\ $G(2^{nd}\ best) > \sqrt{\frac{R^2ln(1/\delta)}{2n_l}}$ or $\sqrt{\frac{R^2ln(1/\delta)}{2n_l}} < \tau $}{
				Split leaf $l$ on the best attr.\\
				Initialize count $n_{ijk}$ for each leaf\\
			}
		}
	}
\end{algorithm}

\setlength{\textfloatsep}{0.5mm}
\setlength{\floatsep}{0.5mm}

\LinesNumbered
\begin{algorithm}[t]
	\footnotesize
	\caption{Incremental Gaussian approximation}
	\label{alg: gausapp}
	\SetKwInOut{Input}{input}
	\SetKwInOut{Output}{output}
	\Input{samples denoted as $(attr_{val}, weight)$}
	\Output{mean of Gaussian approximation denoted as $M$}
	\Output{variance of Gaussian approximation denoted as $V$}
	weight sum: $w\_sum \leftarrow first\ weight$ \\
	variance sum: $v\_sum \leftarrow 0$ \\
	$M \leftarrow first\ attr_{val}$ \\
	\For{each sample $(attr_{val}, weight)$ in sample set}{
		$w\_sum \leftarrow w\_sum + weight$ \\
		$M_{prior} \leftarrow M$ \\
		$M \leftarrow M + \frac{attr_{val} - M_{prior}}{w\_sum}$ \\
		$v\_sum \leftarrow v\_sum + (attr_{val} - M_{prior}) \times (attr_{val} - M)$ \\
		$V \leftarrow \frac{v\_sum}{w\_sum-1}$ \\
	}
\end{algorithm}
\setlength{\textfloatsep}{0.5mm}
\setlength{\floatsep}{0.5mm}

\textbf{2. High Computational Intensity with Dependency and Long Latency.} To reduce memory utilization, Gaussian-based methods~\cite{ga08, streamdm} are applied to trade much higher computational intensity for memory efficiency. For each numeric attribute per class, the sample distribution is estimated in a form of Gaussian distribution. As the Gaussian function is determined by only two values, namely, mean and variance, the memory usage can be significantly compressed to $\# attribute \times \# class \times 2$ per node. However, the incremental update process of the mean and variance leads to high computational demand, as shown in Algorithm~\ref{alg: gausapp}. The requirement of computation resources is proportional to both the number of attributes and classes. Besides this, the split judgment stage also requires computing the cumulative density functions (CDFs) at each split point, which entails even higher computational power. Moreover, the update process incurs long latency and should be in order of time if the two successive iterations work on the same label. In addition to the high computational intensity, the long latency and data dependency further hinder this method from being effectively optimized in hardware.

\vspace{-3mm}
\section{Methodology}
As BRAM and DSP are limited resources for FPGAs, the excessive use of either on-chip memory or computation units in the aforementioned methods~\cite{vfml, vfdtc, ga08, streamdm} is neither efficient nor scalable while handling numeric attributes. The two design challenges described above and their interplay should be taken into consideration for joint optimization. To this end, we propose to introduce an up-to-date quantile algorithm in the induction of online decision trees. 

\LinesNotNumbered
\begin{algorithm}[t]
	\footnotesize
	\caption{Hoeffding tree induction with quantiles}
	\label{alg: qtht}
	\SetKwInOut{Input}{input}
	\SetKwInOut{Output}{output}
	\Input{streaming samples denoted as $(x, y)$}
	\Output{Hoeffding tree structure denoted as $HT$}
	Let $a_i\ (1 \leq i \leq |A|)$ denote the attribute in set A \\
	Let $c_j\ (1 \leq j \leq |C|)$ denote the class in set C \\
	Let ${\alpha}_k (1 \leq k \leq |Q|)$ denote the quantile index \\
	\setcounter{AlgoLine}{0}
	\nl \For{each $(x_t, y_t)$ $\in$ sample set}{
		\nl filter $(x_t, y_t)$ to leaf $f$ of $HT$ \\
		\nl sample num. at $f$: $n_f \leftarrow n_f+1$ \\
		\nl \For{j from 1 to $|C|$}{
			\nl  sample num. in class $j$: $n_{fj} \leftarrow (y_t ==j)\ ?\ n_{fj}+1 : n_{fj}$ \\
		}
		\nl \For{i from 1 to $|A|$}{
			\nl max. attr. value: $max_{a_i} \leftarrow (a_i > max_{a_i}) \ ? \ a_i\ :\ max_{a_i}$ \\
			\nl min. attr. value: $min_{a_i} \leftarrow (a_i < min_{a_i}) \ ? \ a_i\ :\ min_{a_i}$ \\
			\nl \For{j from 1 to $|C|$}{
				\nl \If{$y_t == j$}{
					\nl \For{k from 1 to $|Q|$}{
						\nl ${Q}_{ijt}({\alpha}_k) \leftarrow {Q}_{ij{t-1}}({\alpha}_k) - \lambda sgn_{\alpha}({Q}_{ij{t-1}}({\alpha}_k)-a_i)$ \\
					}
				}
			}
		}
		\nl \If{split trial is activated}{
			\nl \For{i from 1 to $|A|$}{
				\nl \For{p from 1 to $|P|$}{
					\nl $pt \leftarrow \frac{max_{a_i}-min_{a_i}}{|P|+1}\times p + min_{a_i}$ \\
					\nl \For{j from 1 to $|C|$}{
						\nl left distribution $L$: $dist_{Lij}(pt) \leftarrow 0$ \\
						\nl \For{k from 1 to $|Q|$}{
							\nl $dist_{Lij}(pt) \leftarrow (pt > {Q}_{ijt}({\alpha}_k))\  ?\ dist_{Lij}(pt) + 1 : dist_{Lij}(pt)$ \\
						}
						\nl  $dist_{Lij}(pt) \leftarrow \frac{dist_{Lij}(pt)}{|P|} \times n_{fj} $\\
						\nl $dist_{Rij}(pt) \leftarrow n_{fj} - dist_{Lij}(pt)$ \\
					}
				}
				\nl compute $G(a_i)$ for all $pt$ \\
			}
			\nl\If{$G(best)$-$G(2^{nd}\ best) > \sqrt{\frac{R^2ln(1/\delta)}{2n_f}}$ or $\sqrt{\frac{R^2ln(1/\delta)}{2n_f}} < \tau $}{
				\nl split $l$ on the best attr \& initialize new leaves\\
			}
		}
	}
\end{algorithm}

\subsection{Quantile Estimation Using Asymmetric Signum Functions}
Quantiles~\cite{rob96} are cutting points dividing the range of a probability distribution into a certain number of intervals with equal probabilities. The quantile function $Q(\cdot)$ of a continuous variable is defined as the inverse of the CDF, $F(z) = Pr(x_t \leq z)$. Specifically, $Q(\cdot)$ can be written as
\begin{equation}
	\small
	\label{eq:qt}
	Q(\alpha)=F_X^{-1}(\alpha)=inf\{x\in supp(F_X):\alpha\leq F_X(x)\}.
\end{equation}

The state-of-the-art quantile estimation using asymmetric signum functions is studied in~\cite{qt11} and~\cite{qt17}. The quantile approximation calibrates the quantiles in a sequential manner according to every incoming sample. The quantile calibration process from sample $x_{t-1}$ to $x_t$ can be described as
\begin{equation}
	\small
	\label{eq: qtupdate}
	Q_t(\alpha) =  Q_{t-1}(\alpha) - \lambda sgn_{\alpha}(Q_{t-1}(\alpha)-x_t),
\end{equation}
where $sgn_{\alpha}(\cdot)$ is the asymmetric signum function defined by
\begin{equation}
	\small
	\label{eq:sn}
	sgn_{\alpha}(z) = \begin{cases}
		\alpha, \quad if\ z < 0\\
		1- \alpha, \quad if\ z \geq 0
	\end{cases}.
\end{equation}

\subsection{Learning Numeric Attributes with Quantile Approximation}
\begin{figure}[t]
	\begin{center}
		\includegraphics[width=0.7\linewidth]{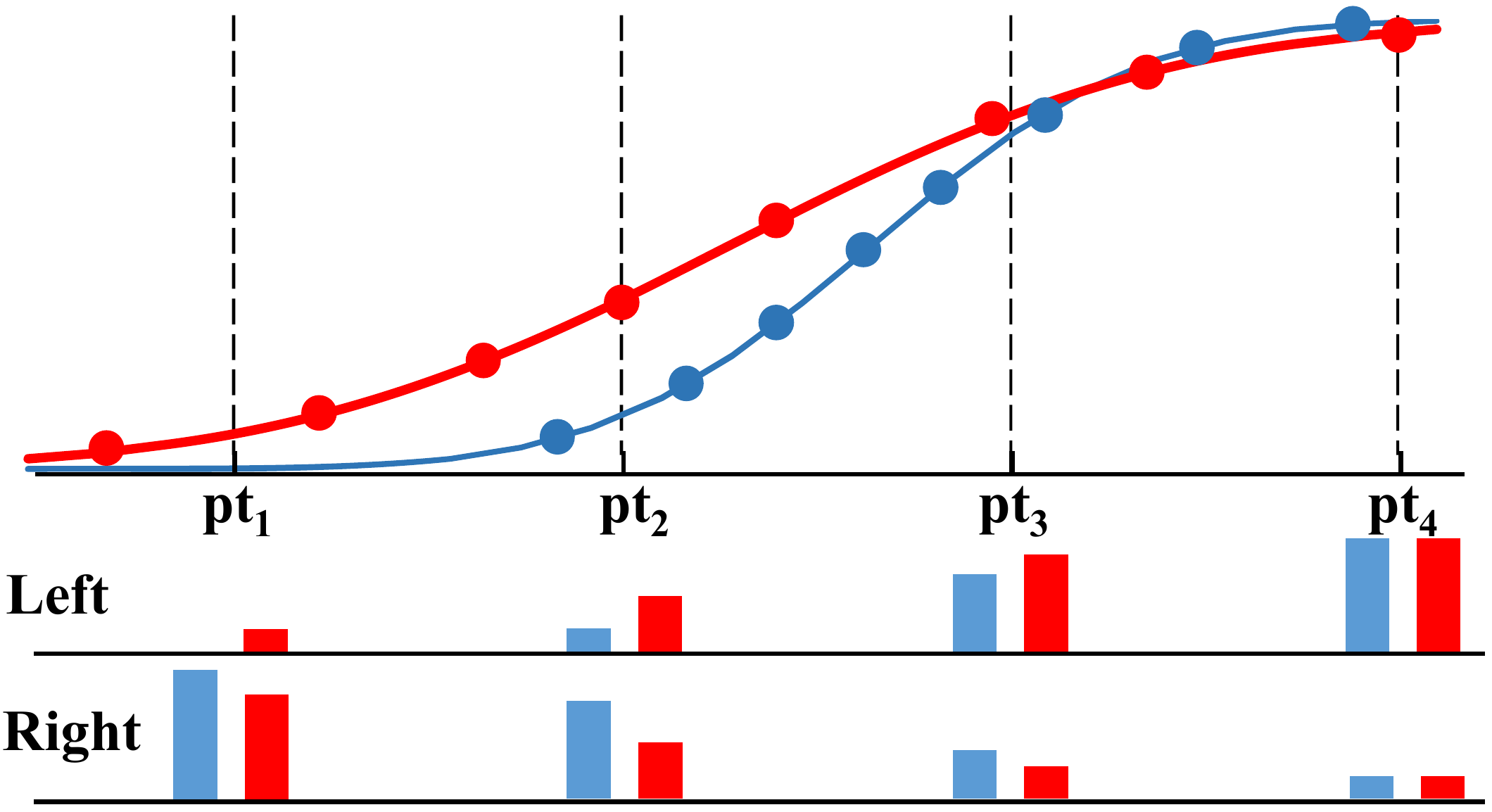}
		\vspace{-3mm}
		\caption{Partition strategy in the proposed algorithm, illustrated with one attribute, two labels and eight quantiles.}
		\label{fig: dist}
	\end{center}
	\vspace{-1mm}
\end{figure}

\begin{figure}[t]
	\begin{center}
		\includegraphics[width=0.65\linewidth]{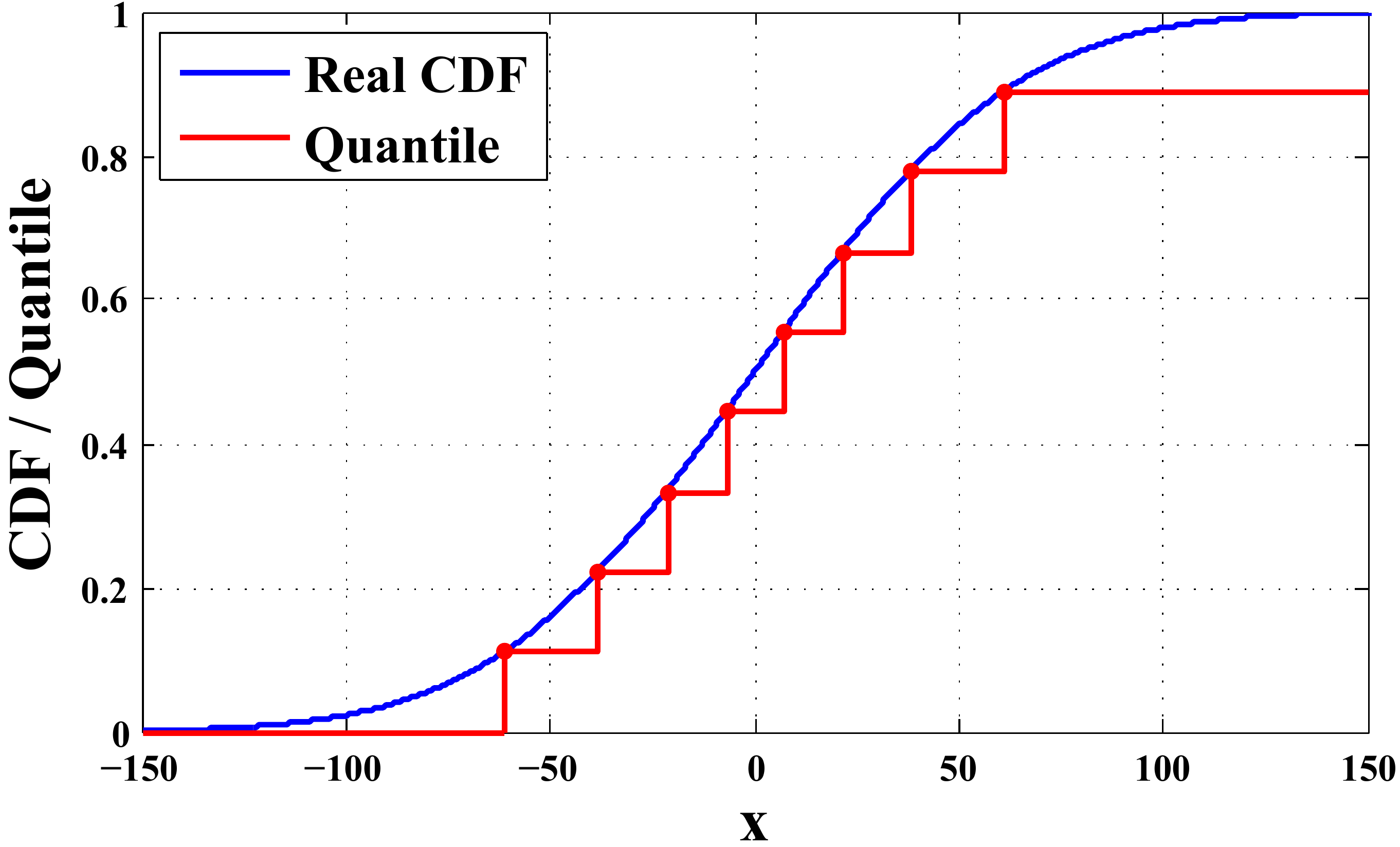}
		\vspace{-3mm}
		\caption{Using eight quantiles to estimate the CDF of normal distribution with a round-down scheme.}
		\label{fig: qt}
	\end{center}
	\vspace{-2mm}
\end{figure}

To handle numeric attributes, we develop a new algorithm in the Hoeffding tree induction process by applying the quantile estimation with asymmetric signum functions, which is described in Algorithm~\ref{alg: qtht}. The proposed algorithm encompasses two key features: 1) a separate set of quantiles is maintained per attribute per class (line 6 to 12); and 2) the strategy to get left/right partitions based on the attribute distributions (line 14 to 22) has been customized to support the quantile method. Note that the number of quantiles to use is determined by the characteristics of the datasets. This is studied in Section~\ref{subsec: qtnum}.

A straightforward method~\cite{vfml} to deduce the partitions is to view each sample as a split point and compute distribution individually: for an attribute $i$ and a specific sample's attribute as the split point $pt_i$, an arbitrary sample is sorted to the left partition if its attribute value $a_i \leq pt_i$, or otherwise, it is filtered to the right partition. In our algorithm, we learn the samples with quantiles and represent sample distribution in CDF: each quantile value $Q({\alpha}_k)$ indicates that the percentage is ${\alpha}_k$ for the samples with the attribute values smaller than $Q({\alpha}_k)$. In this way, sample storage is not required. 

Fig.~\ref{fig: dist} illustrates how the overall partitioning strategy works. We generate a set of split points evenly distributed in the full range of attribute values. These split points are compared to the quantiles individually to find out the interval of two quantiles [$Q({\alpha}_k)$, $Q({\alpha}_{k+1})$] containing the split point. Afterwards, the sample number in each partition can be determined. The portion of samples with attribute values smaller than or equal to $Q({\alpha}_k)$ goes to the left partition, whereas the others go to the right partition. By this method, the sample distribution in the left partition is rounded down to the nearest quantile, with an example shown in Fig.~\ref{fig: qt}.

The proposed algorithm overcomes the trade-off between memory and computation, and presents a more rational and balanced solution compared with state-of-the-art methods~\cite{vfml, vfdtc, ga08, streamdm}. The advantages of this proposed method are three-fold. Firstly, the sample characteristics are fully generalized and encapsulated in a set of quantiles, dispensing with the need to store any samples in the training iterations. The memory requirement is reduced to $\#attribute \times \#label \times \#quantile$ per leaf node. This outperforms existing methods~\cite{vfml, vfdtc} which require large attribute or sample storage. Secondly, the computation demand is notably reduced compared with the memory-efficient yet computation-intensive method, Gaussian method~\cite{ga08, streamdm}: only comparison and subtraction are involved in quantile approximation, whereas Gaussian approximation entails expensive computation as shown in Algorithm~\ref{alg: gausapp}. The complexity of partition deduction is also effectively simplified with the proposed method. Thirdly, the problem of data dependency can be resolved with hardware optimization through deliberate parallelism and pipelining, as introduced in Section~\ref{subsec: lna}.

\section{Architecture Design}
\label{sec: impl}
\subsection{System Overview}
\begin{figure}[t]
	\begin{center}
		\includegraphics[width=0.6\linewidth]{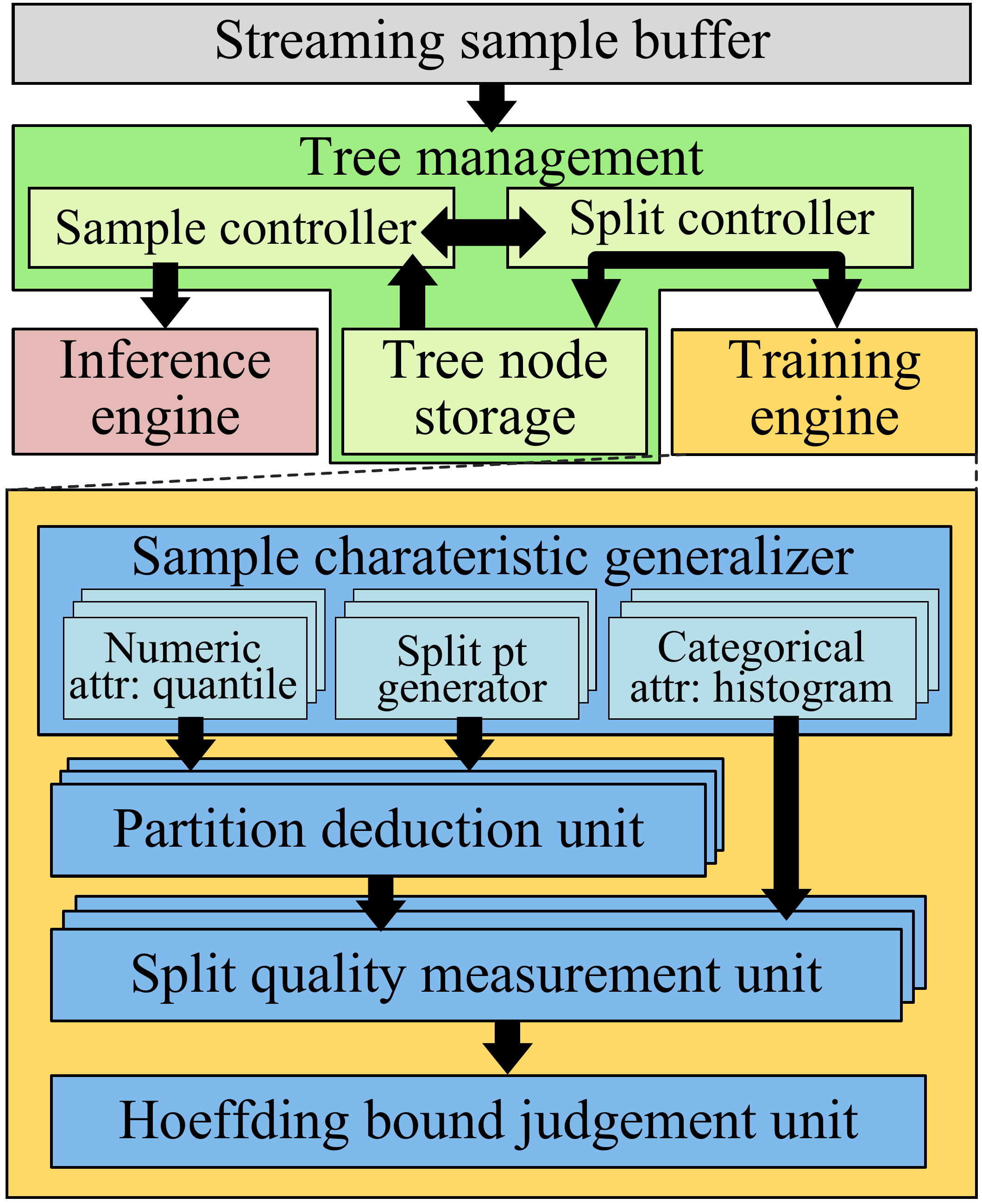}
		\vspace{-2mm}
		\caption{System overview of the Hoeffding tree implementation with the proposed algorithm.}
		\vspace{-2mm}
		\label{fig: arch}
	\end{center}
\end{figure}

The system overview of the Hoeffding tree implementation is depicted in Fig.~\ref{fig: arch}. Starting from the sample buffer, the tree management engine first reads and decodes the sample information. At the same time, it fetches relevant tree nodes from the tree node storage and filters the samples to the leaf nodes in a pipelined way. Thereafter, both the inference engine and training engine start processing the samples.

In the learning process, samples are decomposed into separate attributes and the characteristics of each attribute are learned and stored independently. When a split trial is invoked at a leaf node, for each numeric attribute, a partition deduction unit uses the quantiles and split points to deduce left and right partitions. As for categorical attributes, the sample counts of all attribute-class combinations form a histogram, which is similar to the quantiles for numeric attributes. 

The partition information of every attribute is then processed by a split quality measurement unit to compute the split gain for each split point. Then, the best and second-best split gains are identified, and the split decision is given by the Hoeffding bound judgment unit. If a split is taken, the split information is sent back to the split controller to update the tree structure.

\subsection{Tree Management Units}
\begin{figure}[t]
	\begin{center}
		\includegraphics[width=0.8\linewidth]{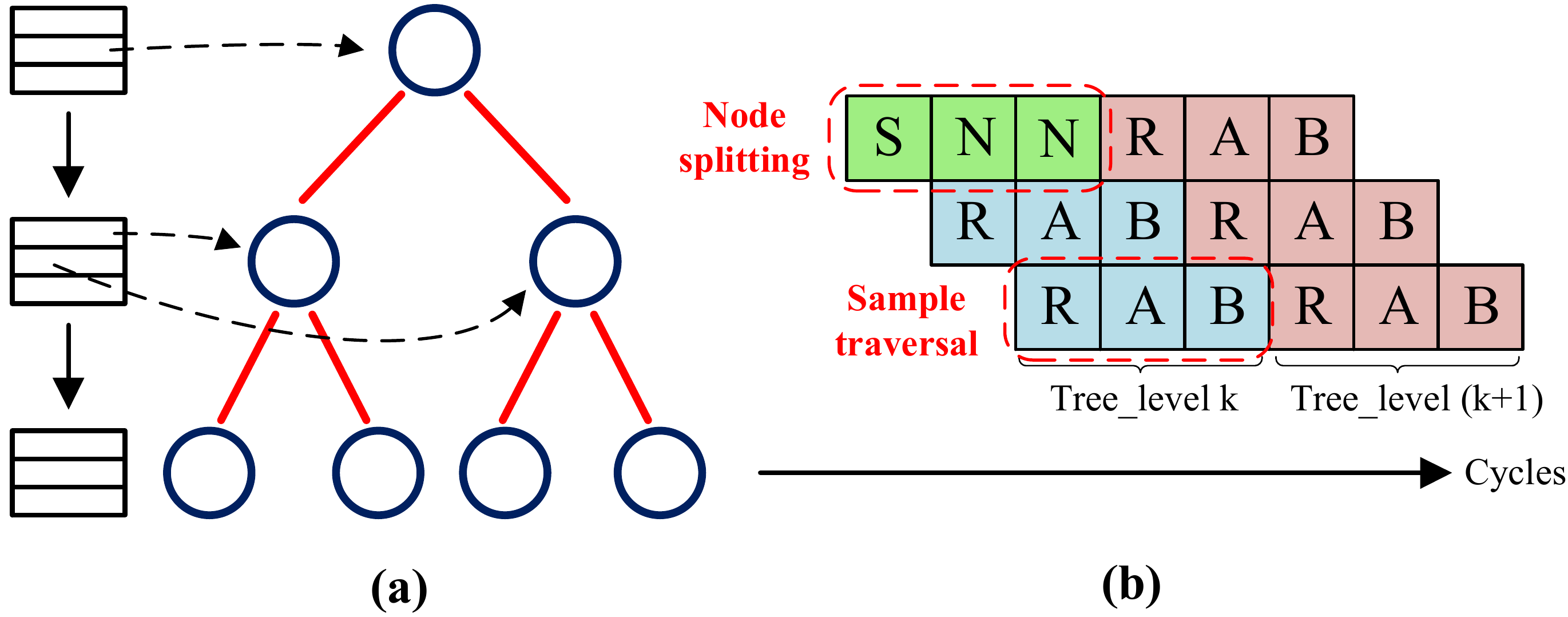}\\
		\includegraphics[width=0.7\linewidth]{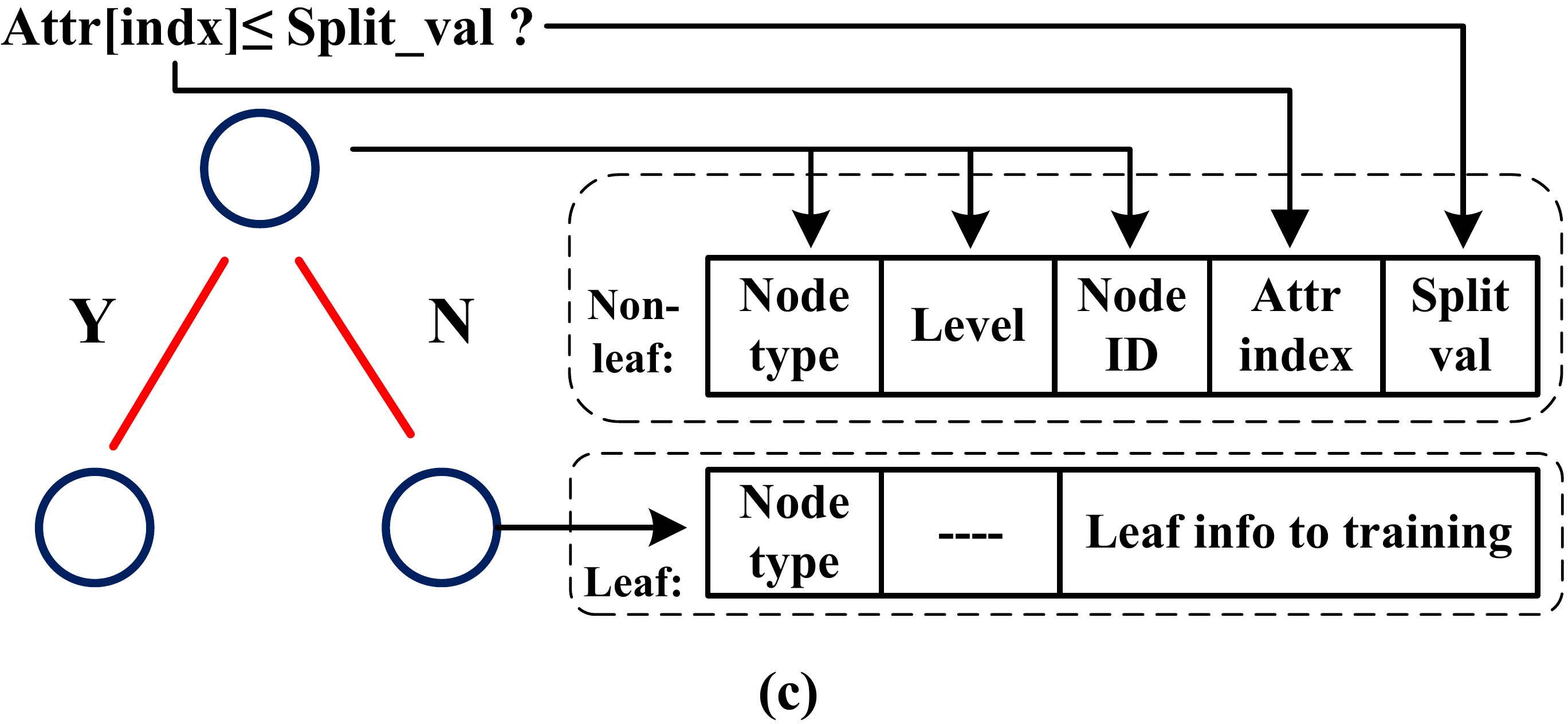}
		\vspace{-3mm}
		\caption{(a) Decision tree architecture; (b) tree management pipeline structure; and (c) bit decomposition of tree node memory.}
		\label{fig: dt}
	\end{center}
\end{figure}

The tree management units maintain two operations: 1) filtering samples to different leaf nodes, which requires tree traversing; and 2) splitting leaf nodes by overwriting the tree node memory after receiving split requests.

The tree traversing process for each sample starts from the root node down to a specific leaf node, thus involving several rounds of memory reading. Considering the case of streaming data input, the tree memory may receive multiple read requests from different samples concurrently. Multi-port memory can be used to support this feature. However, the required port number is linearly related to the tree depths. FPGA BRAMs naturally support up to two ports, and increasing the port size turns out to be an inefficient solution. We observe that the samples are processed at different tree levels sequentially and the samples from different time steps require memory reading from different tree levels. Hence, we separate the node storage according to tree levels, as depicted in Fig.~\ref{fig: dt} (a), and dual-port memory is enough to support both node splitting and tree traversing for streaming samples. The idea of using a separate memory structure has been adopted in DT-CAIF~\cite{saq15}, whereas we develop a fine-grained pipeline structure for each tree level. All the tree levels together form a deep pipeline.

The fine-grained pipeline needs to support both tree traversing and node splitting. A three-stage pipeline is formed, as shown in Fig.~\ref{fig: dt} (b). The tree traversing routine consists of node reading (R), attribute selection (A) and branch decision (B) stages. As for node splitting, split information (mainly the split node level, node ID, split coefficient and attribute index) from the training engine is passed across different tree levels. When a leaf node is reached, the corresponding memory element is overwritten by the split information to replace the leaf node with an internal node. Moreover, two new leaf nodes are generated in the next level and the split pipeline also writes in the new leaf nodes the training elements they are associated to. This is related to the dynamic leaf node-element allocation scheme discussed in Section~\ref{subsec: lna}. All the operations relevant to the split are completed in the split (S) stage, after which two nop (N) states are followed. The bit information stored in the memory for branch decision is shown in Fig.~\ref{fig: dt} (c). 

\subsection{Learning Numeric Attributes}
\label{subsec: lna}
In our proposed Algorithm~\ref{alg: qtht}, recall that we maintain a set of quantile values per numeric attribute per class for a single leaf node. Optimization techniques are investigated for accelerating quantile learning from the hardware perspective, which can be summarized as: 1) attribute-level (coarse-grained) and quantile-level (fine-grained) parallelism; 2) dynamic and memory-based resource sharing; and 3) pipelining with data forwarding for data dependency removal.

\begin{figure}[t]
	\begin{center}
		\includegraphics[width=\linewidth]{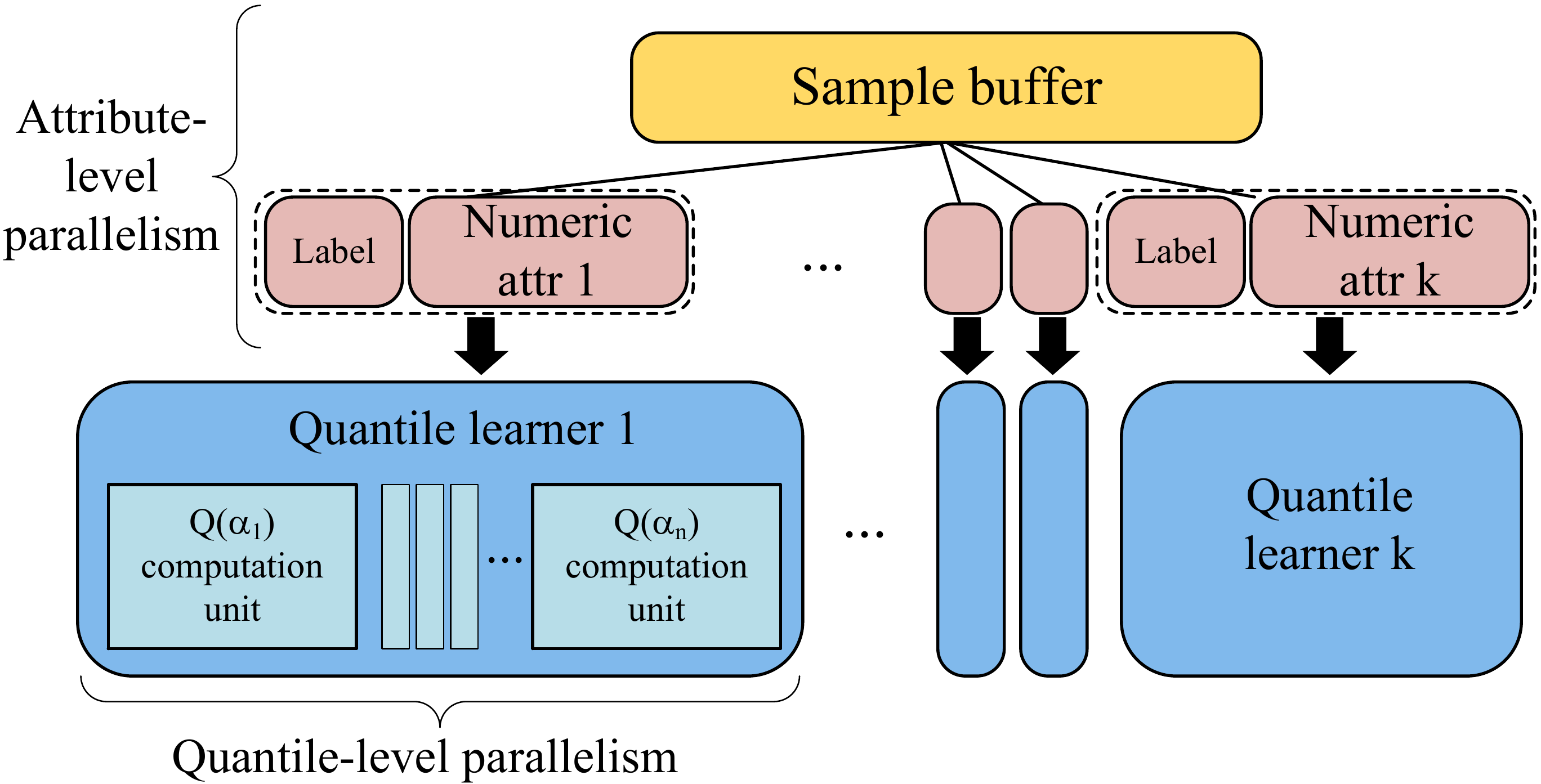}
		\vspace{-7mm}
		\caption{Exploiting attribute-level and quantile-level parallelism.}
		\label{fig: qtpar}
	\end{center}
\end{figure}

\textbf{Attribute-level (Coarse-grained) and Quantile-level (Fine-grained) Parallelism.} As shown in line 6 to line 12 of Algorithm~\ref{alg: qtht}, different attributes are independent and, within each attribute, the quantiles $Q(\cdot)$ per class are also independent of each other. This allows us to speed up the quantile computation process with both attribute-level and quantile-level parallelism, as shown in Fig.~\ref{fig: qtpar}. Note that we do not take class-level parallelism even though it is possible. This is because each sample contains a unique class label but has multiple attributes. The learning process only needs to update the set of quantiles matching the sample label. Based on this fact, parallelizing at class level does not offer any benefit. Instead, we seek opportunities for class-level optimization through resource sharing and pipelining.

\textbf{Dynamic and Memory-based Resource Sharing.} For each leaf node, it is required to maintain a number of quantiles per attribute per class. If hardware copies are simply replicated for each leaf node, both the memory and arithmetic resource utilization becomes too expensive for hardware to implement. In light of this problem, we develop a dynamic leaf node-element allocation scheme as the tree grows dynamically and a memory-based resource sharing mechanism for quantile update routine. 

\begin{figure}[t]
	\begin{center}
		\includegraphics[width=\linewidth]{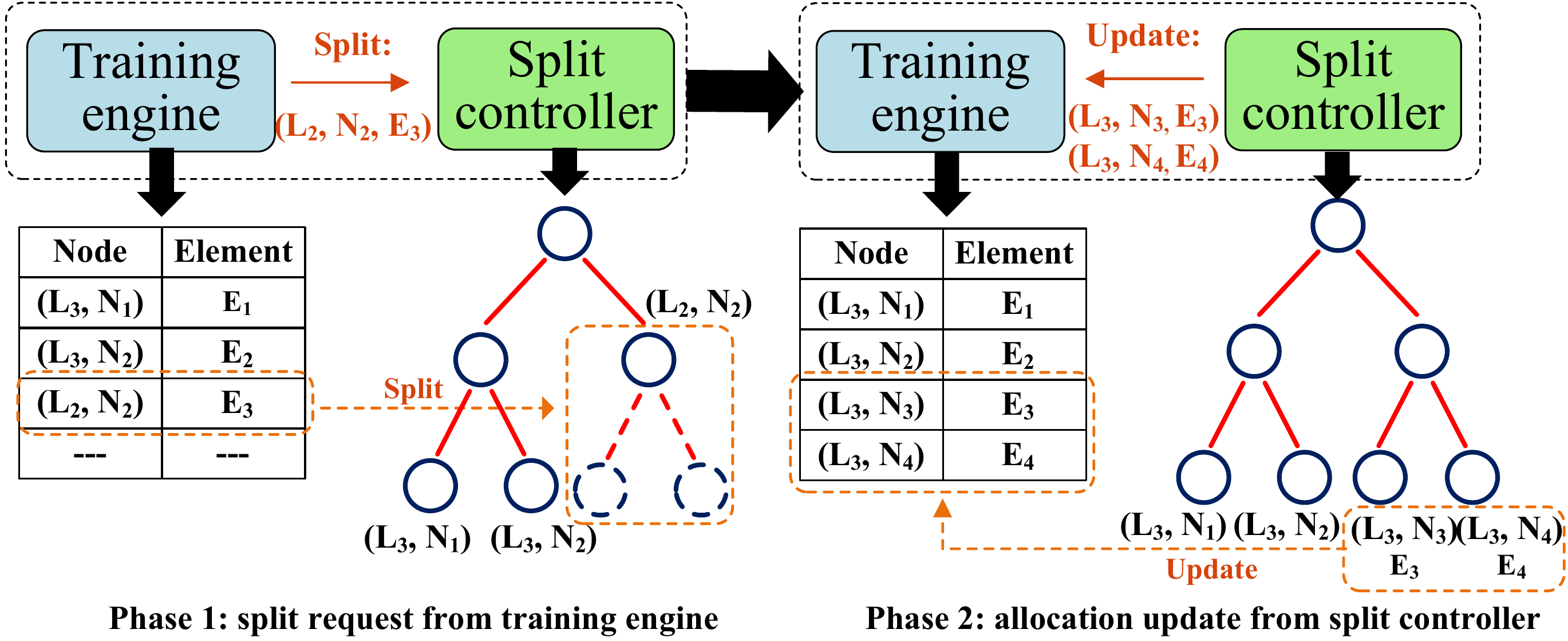}
		\vspace{-7mm}
		\caption{Dynamic leaf node-element allocation scheme.}
		\label{fig: dyn}
	\end{center}
	\vspace{-1mm}
\end{figure}

To differentiate between a leaf node of the tree and the physical resource allocated for a leaf node in the training process, we call the former a \emph{leaf node}, while we denote the latter as an \emph{element}. A leaf node is only temporarily being a leaf node, and it may be split as samples assemble. Therefore, it is not necessary to statically allocate physical resources to each leaf node. We devise a \emph{dynamic leaf node-element allocation scheme}, as shown in Fig.~\ref{fig: dyn}. The training engine maintains a node-element table to keep track of the leaf node-element pairs. During the split process, the split controller generates new leaf node-element pairs and sends them back to the training engine. The traning engine then updates the leaf node-element relationship in the table. In this way, the leaf node-element allocation change dynamically and resource reuse in hardware is facilitated.

\begin{figure}[t]
	\begin{center}
		\includegraphics[width=0.7\linewidth]{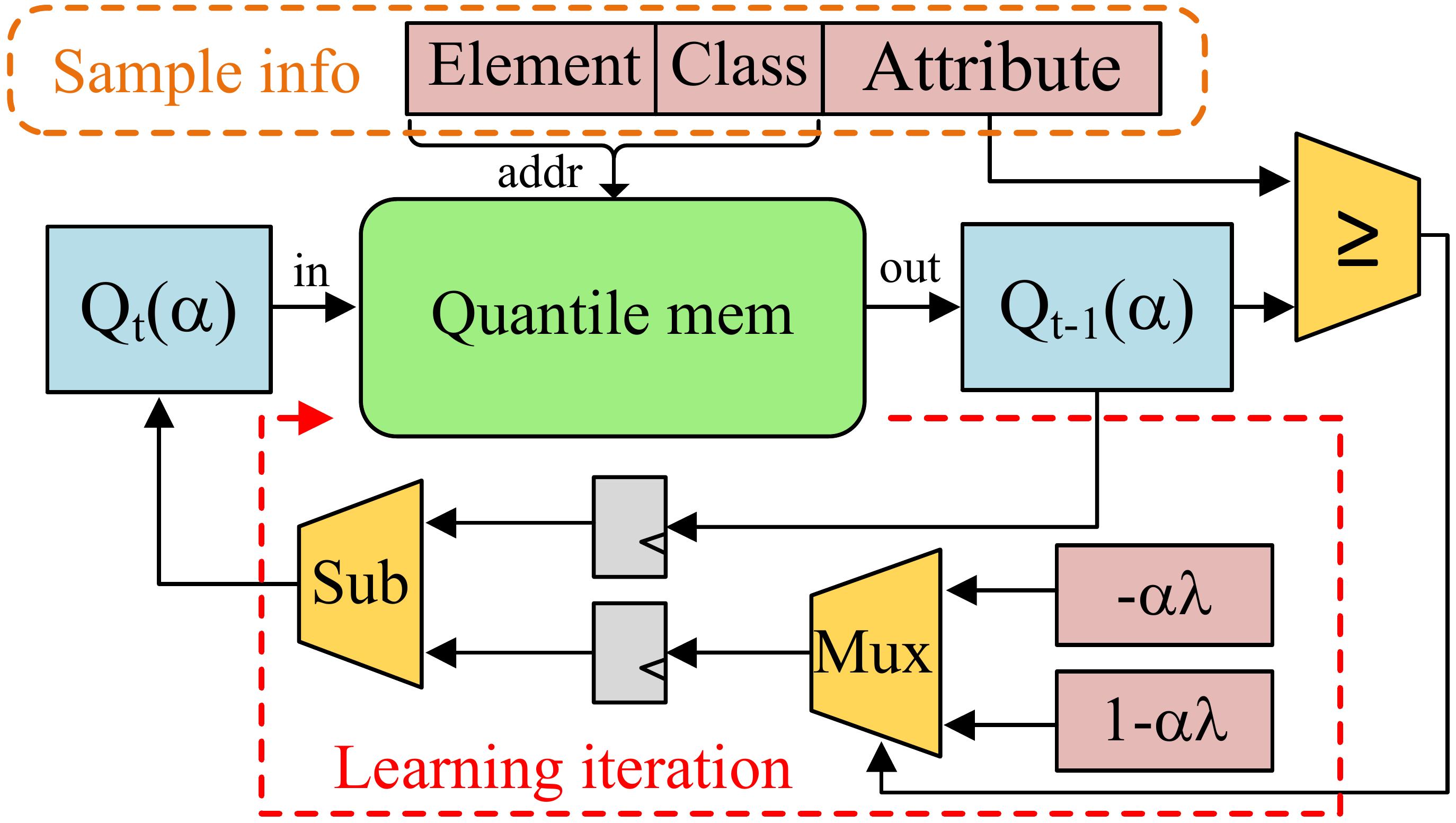}
		\vspace{-2mm}
		\caption{A single quantile computation unit with the memory-based resource sharing scheme.}
		\label{fig: mbqta}
		\vspace{-2mm}
	\end{center}
\end{figure}

A \emph{memory-based resource sharing scheme} is designed to collaboratively work with the dynamic leaf node-element allocation scheme for further resource sharing. This scheme leverages two facts: 1) each sample is only sorted to one leaf node, so only one element will be activated for quantile update per sample; and 2) for each attribute, only the set of quantiles corresponding to the sample label will be activated per sample. Since the quantile learning process is the same for all classes and elements, except that the quantile values are different, we devise the following memory-based resource sharing scheme: for each attribute, all the classes of all elements share one set of quantile computation logics and all the corresponding quantile values are stored in one memory. When a sample is used for training, the set of quantiles corresponding to the specific element and class is fetched, and later, the updated values are stored back to the same memory location. Element and class values together form the memory addresses. Putting it all together, a single quantile computation unit with memory-based resource sharing is depicted in Fig.~\ref{fig: mbqta}. To support this mechanism, each leaf node in the tree memory preserves a field denoted as \emph{leaf information to training} shown in Fig.~\ref{fig: dt} (c). Provided a new split, the two new leaf nodes along with their assigned element IDs are sent from the training engine to the tree node memory for update. For each sample after tree traversing, the element ID associated with its reached leaf node and the raw sample data are sent to the training engine.

\textbf{Pipelining with Data Forwarding.} There exists data dependency for quantile computation: two successive samples sorted to the same leaf node should update the same element in a sequential way. For the method with Gaussian approximation, the long latency of the update process, as described in Algorithm~\ref{alg: gausapp}, makes it difficult to overcome this dependency. For the proposed quantile computation architecture in Fig.~\ref{fig: mbqta}, the computation is reduced to a comparison and a subtraction per quantile unit, which allows us to fully exploit the pipeline architecture with data forwarding to resolve data dependency. 

\begin{figure}[t]
	\begin{center}
		\includegraphics[width=\linewidth]{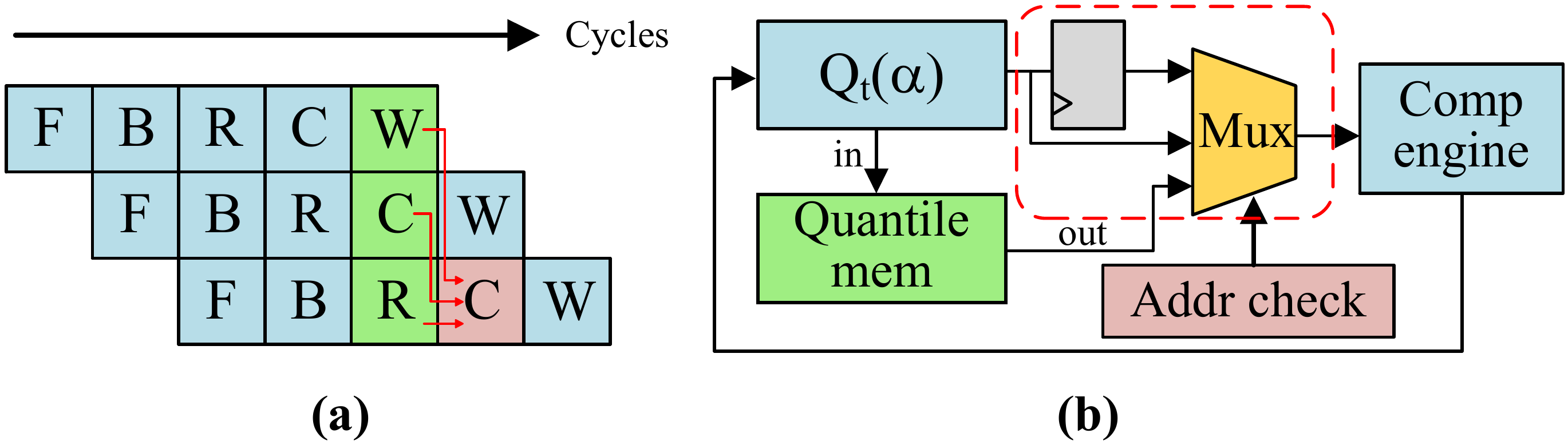}
		\vspace{-7mm}
		\caption{(a) Pipelining stages with data forwarding; (b) hardware realization of data forwarding.}
		\label{fig: qtpipe}
	\end{center}
	\vspace{-2mm}
\end{figure}

We propose a 5-stage pipeline architecture for the quantile update routine, as shown in Fig.~\ref{fig: qtpipe} (a). The first stage (F) fetches a sample from the sample buffer. The second stage (B) decides on the execution branch to take, including element initialization in the dynamic leaf node-element allocation scheme, quantile computation and quantile output for the split trial. In the next stage (R), the quantile unit selected by the element and class is read out. Afterwards, the quantile is updated in the computation stage (C) following Equation~(\ref{eq: qtupdate}), and is written back to the same memory location in the writing stage (W).

In stage C, we address the data dependency problem by the adoption of a dedicated data forwarding method, as shown in Fig.~\ref{fig: qtpipe} (b), which aims at providing the flexibility that, when the quantiles are updated while not yet written in the memory, they are directly passed to the quantile computation engine if the addresses between these two computation periods match. We keep track of the results and quantile memory addresses of the prior two computation periods, which are managed by stage C and stage W, respectively. Stage C has a higher forwarding priority over stage W when both memory addresses match the one currently processing, because stage C provides the most up-to-date results. This data forwarding allows us to bypass memory operations when dependency occurs and eventually leads to a throughput of one sample per cycle.

\subsection{Learning Categorical Attributes}
The process of learning categorical attributes is similar to learning numeric attributes. 
However, the value and size of each categorical attribute is determined by dataset characteristics, which can be known in design time. Therefore, counting the number of occurrence for each attribute-class combination gives a histogram of the distribution without any loss of information. In a split trial for categorical attributes, each attribute value is used as a split point individually: the samples with the attribute value equal to the split point is filtered to the left, or otherwise, it is sorted to the right.

\begin{figure}[t]
	\begin{center}
		\includegraphics[width=0.9\linewidth]{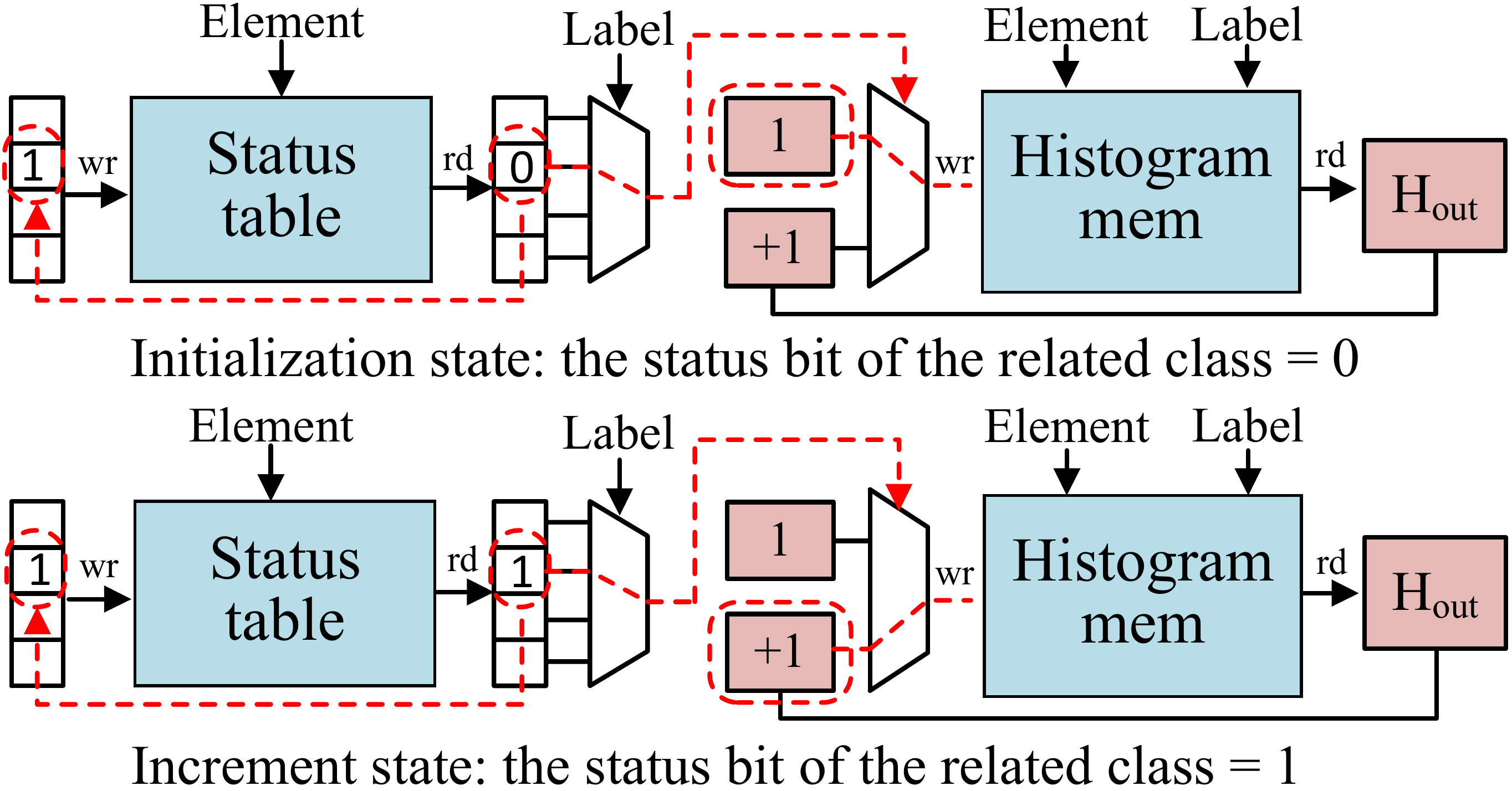}
		\vspace{-3mm}
		\caption{Histogram update with status table.}
		\label{fig: hist}
		\vspace{-2mm}
	\end{center}
\end{figure}

The optimization methods, except the dynamic leaf node-element allocation scheme, can be migrated to categorical attributes seamlessly. However, to support the dynamic leaf node-element allocation scheme, the histograms of all attribute-class combinations for an element need to be initialized simultaneously. This brings difficulties as we apply memory-based resource sharing in which the same dual-port histogram memory is shared amongst different class labels, and multiple write requests to the same memory is inefficient for FPGA design. To overcome this problem, we additionally implement a status table for histograms. Every memory unit in the status table represents an individual histogram, and each bit indicates the status of a column of this histogram. 
To initialize a histogram when a new leaf node-element pair is assigned, only the corresponding memory unit in the status table, instead of all units in the histogram, needs to be reset. The training routine first checks the status table for each incoming sample, and follows either of the two situations (i.e., initialization or increment) as depicted in Fig.~\ref{fig: hist}. The relevant status bit is set to high when the first sample comes after initialization.

\begin{table*}[t]
	\centering
	\begin{threeparttable}
		\footnotesize
		\caption{Inference accuracy using different numbers of quantiles.}
		\vspace{-7mm}
		\label{table: acc}
		\begin{tabular}[width=\linewidth]{c|c|c|c|c|c|c|c|c|c|c|c}
			\toprule
			\multirow{2}{*}{\textbf{Dataset}} & \multicolumn{1}{c|}{\textbf{Gaussian}} & \multicolumn{9}{c}{\textbf{Quantile method with different quantile size}} \\
			& \multicolumn{1}{c|}{\textbf{method}} & \multicolumn{1}{c|}{2} & \multicolumn{1}{c|}{4} & \multicolumn{1}{c|}{8} & \multicolumn{1}{c|}{16} & \multicolumn{1}{c|}{24} & \multicolumn{1}{c|}{32} & \multicolumn{1}{c|}{64} & \multicolumn{1}{c|}{128} & \multicolumn{1}{c|}{256} & \multicolumn{1}{c}{512}\\
			\midrule
			\multicolumn{1}{c|}{Bank} & 89.10\% & 88.79\% & 89.05\% & 89.15\% & 89.30\% & 89.31\% & 89.32\% & 89.26\% & 88.52\% & 88.66\% & 88.59\% \\
			\multicolumn{1}{c|}{Telescope} & 76.16\% & 76.68\% & 74.61\% & 76.41\% & 76.12\% & 76.15\% & 76.64\% & 75.51\% & 75.75\% & 76.75\% & 71.32\% \\
			\multicolumn{1}{c|}{Electricity} & 76.26\% & 76.97\% & 77.26\% & 78.02\% & 76.31\% & 77.93\% & 77.53\% & 76.91\% & 76.75\% & 76.61\% & 74.15\% \\
			\multicolumn{1}{c|}{Covertype} & 71.02\% & 72.46\% & 72.17\% & 72.72\% & 72.51\% & 72.50\% & 71.86\% & 73.43\% & 71.90\% & 70.94\% & 69.41\%\\
			\multicolumn{1}{c|}{Person} & 39.00\% & 45.90\% & 48.82\% & 51.38\% & 52.49\% & 53.37\% & 52.35\% & 52.40\% & 47.94\% & 47.44\% & 49.60\%\\
			\bottomrule
		\end{tabular}
	\end{threeparttable}
	\vspace{-3mm}
\end{table*}

\subsection{Simplification of Split Measurement with Hoeffding Bound}
The study in~\cite{tan07} has shown that the choice of split measurement method does not exert a significant impact on the accuracy of decision tree induction. We adopt gini impurity~\cite{cart} as it is commonly used and has low computational demand.

Gini impurity is a measure of the chance for an example to be incorrectly classified if it is randomly labeled according to the distribution of the labels. Let $p_j$ be the probability of examples being labeled as class $j\ (j \in{1,2,...,|C|})$ in the dataset $S$. Gini impurity can be represented as
\begin{equation}
	\small
	\label{eq: gini}
	gini(S) = 1-\sum_{j=1}^{|C|}p_j^2.
\end{equation}
The split quality for a given partition is based on the reduction in gini impurity after a split is taken. If $S$ is split into the left subset $S_L$ and right subset $S_R$, the reduction in gini impurity can be described by
\begin{equation}
	\small
	\label{eq: ginigain}
	G = \Delta gini = gini(S) - \frac{|S_L|}{|S|}gini(S_L) - \frac{|S_R|}{|S|}gini(S_R).
\end{equation}

We combine the split measurement with the Hoeffding bound judgment for joint optimization in hardware. Let $S_{L, j}$ and $S_{R, j}$ be the subset of $S_L$ and $S_R$ labeled in $j$, respectively. We reorganize the reduction in gini impurity $G$ as follows:
\begin{equation}
	\small
	\label{eq: reginigain}
	G = \frac{1}{|S|}(\frac{1}{|S_L|}\sum_{j=1}^{|C|}|S_{L,j}|^2 + \frac{1}{|S_R|}\sum_{j=1}^{|C|}|S_{R,j}|^2) + gini(S)-1.
\end{equation}
Putting the gini impurity reduction and Hoeffding bound together, the calculation can be reorganized as
\begin{equation}
	\scriptsize
	\begin{aligned}
		\label{eq: ginihb}
		G_{B_1}-G_{B_2} = & \frac{1}{|S|} \big[ \underbrace{(\frac{1}{|S_{B_1,L}|}\sum_{j=1}^{|C|}|S_{B_1,L,j}|^2 +  \frac{1}{|S_{B_1,R}|}\sum_{j=1}^{|C|}|S_{B_1,R,j}|^2)}_{\textnormal{split quality}} - \\& (\frac{1}{|S_{B_2,L}|}\sum_{j=1}^{|C|}|S_{B_2,L,j}|^2 + \frac{1}{|S_{B_2,R}|}\sum_{j=1}^{|C|}|S_{B_2,R,j}|^2) \big].
	\end{aligned}
\end{equation}
To search for the best and second-best attributes, we only need to compute the \emph{split quality} denoted in Equation~(\ref{eq: ginihb}) for each split point, instead of the full term of gini impurity reduction in Equation~(\ref{eq: reginigain}). After that, the whole term of Equation~(\ref{eq: ginihb}) is computed for Hoeffding bound judgment. This noticeably simplifies the calculation for each split point.

To further optimize the computation, we eliminate the division $\frac{1}{|S|}$ in Equation~(\ref{eq: ginihb}) by pre-storing and looking up the values in memory. The square-sum calculation in the split quality term is realized with a pipelined multiplier-adder tree.

\section{Experiments}
\subsection{Experimental Setup}
In the experiments, we put our main focus on online tree learning. The differences in traditional, batch and online tree learning have been studied in prior works~\cite{hang10, pedro00} and are not elaborated in this paper. We first implement the software version of our proposed algorithm in StreamDM-C++\cite{streamdm}, the state-of-the-art software toolkit supporting the Hoeffding tree. The parameter settings related to the Hoeffding bound are $n_{min}$ = 200, $n_{pt}$ = 10, $\tau$ = 0.05, $\delta$ = $10^{-3}$ and $\lambda$ = 0.01, according to~\cite{pedro00, streamdm} and~\cite{qt17}. The maximum leaf number is 1024, and the maximum tree depth is 15. We use a 32-bit fixed-point data representation with a 30-bit fraction for numeric attributes, after normalizing the data to within the range of [-1,1], if necessary. We evaluate the design with five large datasets: Bank Marketing (Bank), MAGIC Gamma Telescope (Telescope), Australian New South Wales Electricity Market (Electricity), Covertype and Person Activity (Person) from the UCI machine learning repository~\cite{uci07} and related works~\cite{streamdm, cheng15}. The optimized hardware is designed in Verilog and implemented on the Xilinx VCU1525 platform~\cite{virtexp} using SDAccel 2018.2. The datasets are transferred from CPU to off-chip memory (DDR4) on the FPGA platform through PCIe.

\vspace{-1mm}
\subsection{Tuning the Number of Quantiles}
\label{subsec: qtnum}
We tune the number of quantiles in a wide range to evaluate the model performance. The evaluation methodology is \emph{Interleaved-test-then-train}: each sample is first passed through testing before it is applied for training. This is a commonly used evaluation method for online learning models, and the model performance is evaluated by inference accuracy for the entire datasets. In this way, both the online training and testing phases fully utilize the whole datasets, which is different from offline training methods that require a train-test division and need to separately evaluate training and testing accuracy. 

Experimental results in Table~\ref{table: acc} show that the inference accuracy may be degraded significantly as the number of quantiles becomes either too small or too large, especially for the Person dataset. When the quantile number is small, the learning ability of the model may be constrained, because the learned distribution is too coarse-grained to provide effective information. Conversely, if the quantile number becomes too large, the generalization ability may be impaired as well, since the design is more prone to noise in the datasets. Setting the quantile number between 8 and 32 provides high accuracy with desirable robustness. Considering the fact that memory and computation demand is proportional to the number of quantiles, we adopt a unified quantile number of 8 in the hardware design. One can also tune the quantile number to best fit a target dataset. Table~\ref{table: resfreq} shows the size of datasets and information about FPGA implementation. 

\begin{table}[t]
	\centering
	\footnotesize
	\begin{threeparttable}
		\caption{Resource utilization and frequency of FPGA design.}
		\vspace{-3mm}
		\label{table: resfreq}
		\begin{tabular}[width=\linewidth]{c|c|c|c|c|c}
			\toprule		
			\multicolumn{1}{c|}{\textbf{Dataset}} & \multicolumn{1}{c|}{\textbf{Size}} & \multicolumn{1}{c|}{\textbf{LUT\tnote{1}}} & \multicolumn{1}{c|}{\textbf{BRAM\tnote{2}}} & \multicolumn{1}{c|}{\textbf{DSP\tnote{3}}} & \multicolumn{1}{c}{\textbf{Freq. (MHz)}}\\
			\midrule
			\multicolumn{1}{c|}{Bank} & 45211 & 63079 & 486 & 202 & 308 \\
			\multicolumn{1}{c|}{Telescope} & 19020 & 73800 & 480 & 184 & 305 \\
			\multicolumn{1}{c|}{Electricity} & 45312 & 54198 & 384 & 138 & 300 \\
			
			\multicolumn{1}{c|}{Covertype} & 581012 & 169334 & 1883 & 1126 & 170 \\
			\multicolumn{1}{c|}{Person} &164860 &  59401 & 986 & 191 & 266 \\
			\bottomrule
		\end{tabular}
		\begin{tablenotes}
			\scriptsize
			\item[1]Total No. LUT: 1182240\item[2]Total No. BRAM: 2160\item[3]Total No. DSP: 6840
		\end{tablenotes}
	\end{threeparttable}
\end{table} 

\begin{table}[t]
	\centering
	\footnotesize
	\caption{Performance comparison: batch learning \& online learning.}
	\label{table: compbatch}
	\begin{tabular}[width=\linewidth]{c|c|c|c}
		\toprule
		\multicolumn{1}{c|}{\textbf{Method}} & \multicolumn{1}{c|}{\textbf{Platform}} & \multicolumn{1}{c|}{\textbf{Freq.}} & \multicolumn{1}{c}{\textbf{Exe. time}}\\
		\midrule
		\multicolumn{1}{c|}{Batch learning~\cite{cheng15}} & Intel Stratix IV  & 200 MHz & 118 s \\
		\multicolumn{1}{c|}{This work} & Xilinx Ultrascale+ & 170 MHz & 3.97 ms \\
		\bottomrule
	\end{tabular}
\end{table} 

\begin{figure*}[t]
	\begin{center}
		\includegraphics[width=5.2cm]{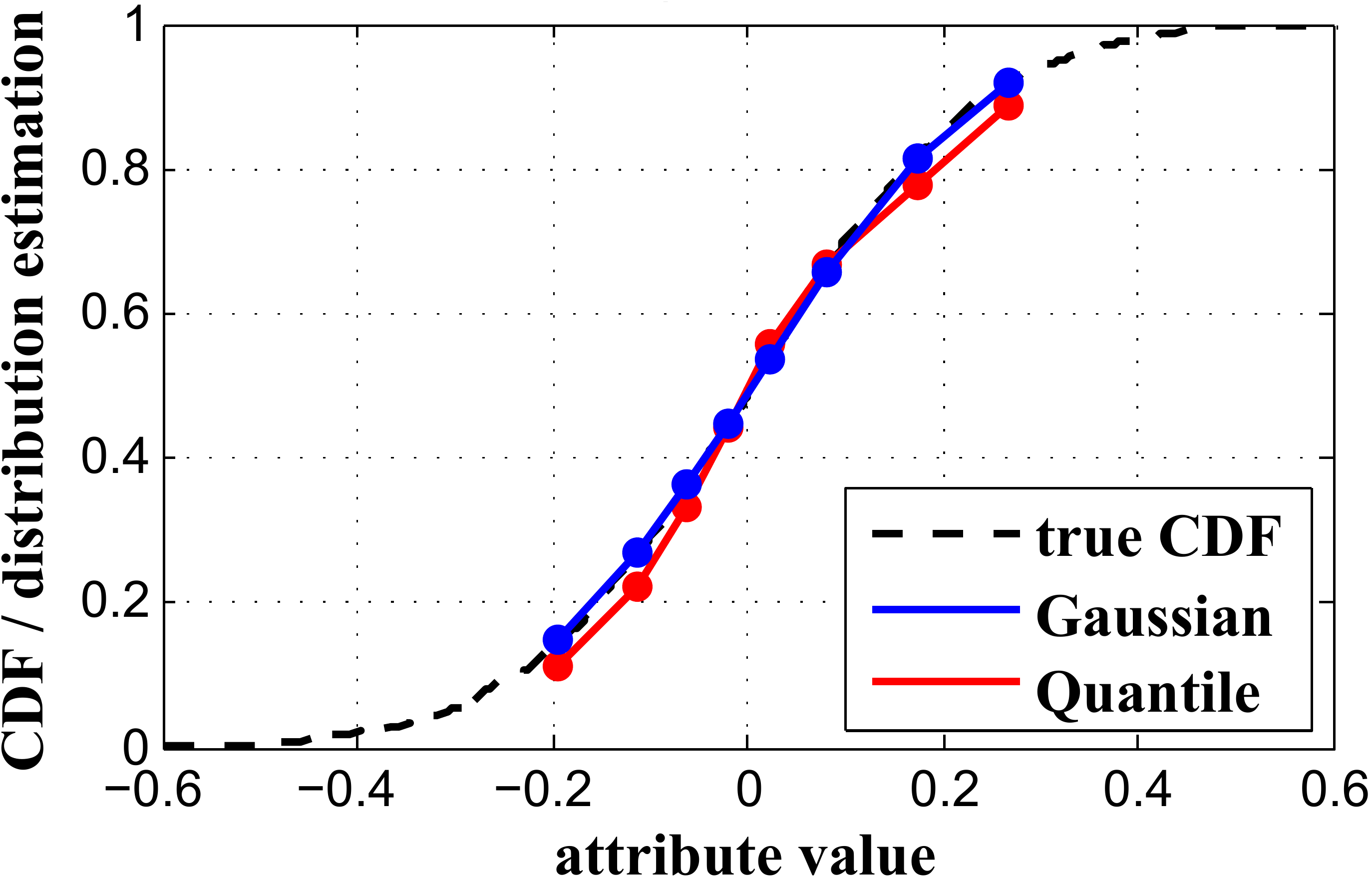}
		\includegraphics[width=5.2cm]{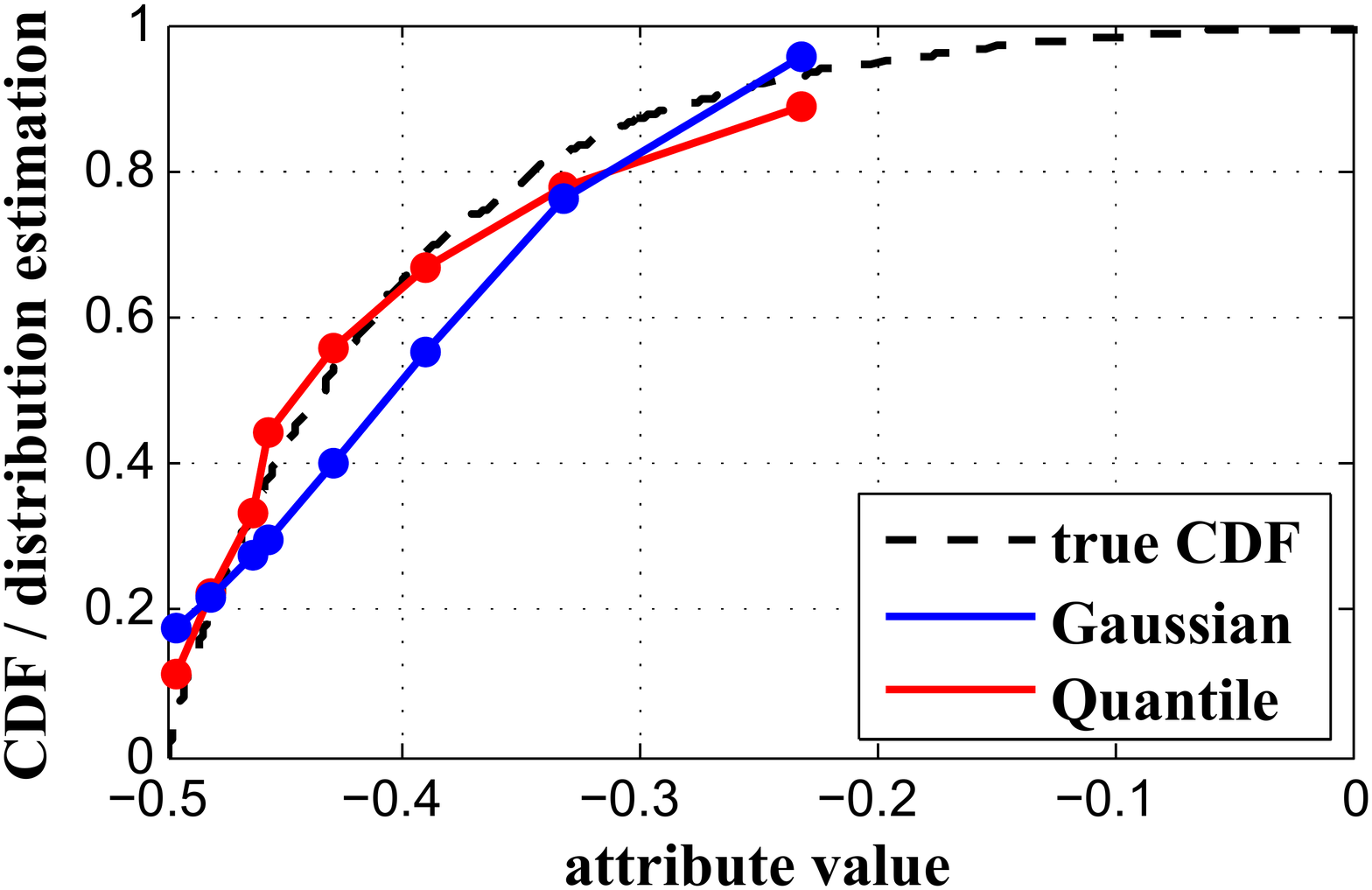}
		\includegraphics[width=5.2cm]{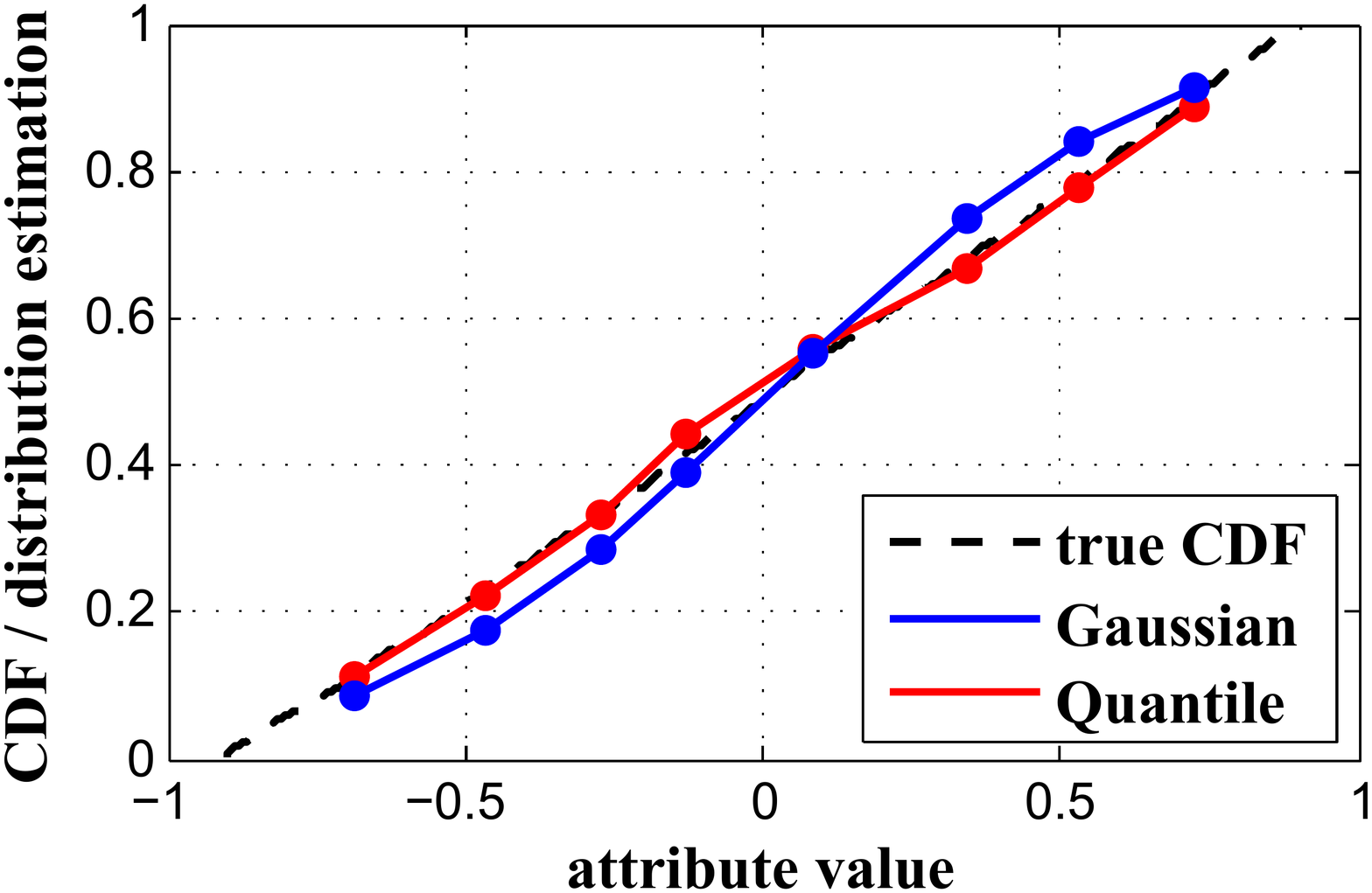}
		\vspace{-2mm}
		\caption{Gaussian and quantile estimation of true CDFs of three representative attributes from Electricity dataset.}
		\label{fig: apprx}
	\end{center}
	\vspace{-8mm}
\end{figure*}

\vspace{-2mm}
\subsection{Comparison with Batch Learning on FPGA}
The up-to-date method to cope with decision tree learning with large datasets on FPGA is through batch learning. The work~\cite{cheng15} presented a state-of-the-art FPGA architecture for batch-based decision trees. Covertype is used as the only benchmark in~\cite{cheng15}, and it serves as the baseline for comparison in Table~\ref{table: compbatch}. The accuracy and overall resource usage are not given, but study in~\cite{pedro00} has proven that both Hoeffding tree and batch tree can lead to the same results for large datasets asymptotically. Table~\ref{table: compbatch} shows that our proposed online learning design can offer an up to 5-orders-of-magnitude speedup in execution time in comparison to~\cite{cheng15}. This significant speedup stems from the difference in communication patterns. The work~\cite{cheng15} involves a number of rounds of transmission for the same samples from and to the off-chip DDR memory in the training process per batch: it reads the sample set at the start of a split process and writes back the subset of samples in each resulting split. By contrast, our proposed online training architecture only requires reading each sample once in the entire learning process, thus reducing a large amount of high-cost inter-chip communication.

\vspace{-2mm}
\subsection{Comparison with the State-of-the-art on Processors}
\begin{table}[t]
	\centering
	\footnotesize
	\caption{Comparison of Software and Hardware Execution Time.}
	\vspace{-2mm}
	\label{table: exetime}
	\begin{tabular}[width=\linewidth]{c|c|c|c|c}
		\toprule
		\multirow{2}{*}{\textbf{Dataset}} & \multicolumn{2}{c|}{\textbf{CPU exe. time}} & \multicolumn{1}{c|}{\textbf{FPGA}}  
		& \multirow{2}{*}{\textbf{Speedup}}\\
		& \multicolumn{1}{c|}{Gaussian} & \multicolumn{1}{c|}{Quantile} & \multicolumn{1}{c|}{\textbf{exe. time}} 
		& \multicolumn{1}{c}{}\\
		\midrule
		\multicolumn{1}{c|}{Bank} & 0.27 s & 0.25 s & 0.36 ms & 750 / 694 $\times$ \\
		\multicolumn{1}{c|}{Telescope} & 0.11 s & 0.10 s & 0.26 ms & 423 / 384 $\times$ \\
		\multicolumn{1}{c|}{Electricity} & 0.21 s & 0.20 s & 0.42 ms & 500 / 476 $\times$ \\
		\multicolumn{1}{c|}{Covertype} & 6.06 s & 6.28 s & 3.97 ms & 1526 / 1581 $\times$ \\
		\multicolumn{1}{c|}{Person} & 0.79 s & 0.75 s & 0.93 ms & 849 / 806 $\times$ \\
		\bottomrule
	\end{tabular}
\end{table}

StreamDM-C++~\cite{streamdm} reported that Gaussian method provided the best performance amongst prior methods~\cite{vfml, vfdtc, ga08, streamdm}, so it is used as the baseline in this paper. Regarding inference accuracy, our proposed algorithm with eight quantiles outperforms the Gaussian method for all five benchmarks, with 0.05\% to 12.3\% improvement, as shown in Table~\ref{table: acc}. 

The results of CDF approximation using the quantile method and Gaussian method account for this gap in accuracy. Three attributes with representative distributions in the Electricity dataset are selected to illustrate the results, as shown in Fig.~\ref{fig: apprx}. The sample set is the subset in the root node before it is split. The CDF of the first attribute is close to the Gaussian function, and thereby, the Gaussian method provides slightly better fitting results than the quantile method. However, regarding the second and third attributes, the quantile method outperforms the Gaussian method. The Gaussian method assumes that the sample distribution conforms with Gaussian distribution, and lead to poor approximation quality for distributions unlike Gaussian. By contrast, the quantile method makes no presumption of any distribution, and hence, it offers accurate approximation for various distributions, including Gaussian distribution. In other words, the quantile method has a wider scope of applicability than the Gaussian method, which accounts for the improvement in accuracy.

For the execution time, we integrate the quantile method in StreadDM-C++ and run this toolkit with both the Gaussian and quantile methods on the Xeon E5-2680 platform under 2.6 GHz. As shown in Table~\ref{table: exetime}, our proposed hardware designs on FPGA achieve 423$\times$ to 1526$\times$ speedup over the Gaussian method and 384$\times$ to 1581$\times$ speedup over the quantile method in software implementation, respectively.

\section{Related Work}
Decision tree acceleration on FPGA has been widely investigated. Most of the existing works~\cite{van12, qu14, owai17, tong17} have targeted FPGA-based acceleration of inference engines. For decision tree training on FPGA, the work~\cite{nara07} migrated the gini computation from software processing to FPGA implementation. The work~\cite{saq15} sought an SoC solution where the training stage was executed by a soft-core processor on FPGA, while the inference unit was implemented with FPGA logics. The work~\cite{cheng13} first designed a complete traditional decision tree training system on FPGA and devised a FIFO-based sorter to facilitate sorting for training, but the memory utilization was high and the maximum size of the datasets was restricted. To reduce memory usage, the work~\cite{cheng14} improved upon~\cite{cheng13} by employing dataset compression and decompression, with additional data preprocessing time. The work~\cite{cheng15} proposed the state-of-the-art of batch learning of decision trees on FPGA. However, it turns out to be inefficient for training large datasets, mainly because it requires transmitting the same samples between the FPGA and off-chip memory multiple times. By contrast, our work introduces a light-weight algorithm along with a hardware-friendly architecture for online decision tree learning, placing no restrictions on the size of datasets and only requiring one-time inter-chip transmission of the datasets. To the best of our knowledge, this paper introduces the first design and optimization of an online decision tree that is applicable to large-scale datasets, and is meanwhile, suitable for FPGA acceleration.

\section{Conclusion}
Online decision tree algorithms suffer from either high memory usage or high computational intensity with dependency and long latency. In this paper, we introduce an efficient and scalable quantile-based induction algorithm for the Hoeffding tree, and we investigate hardware optimization techniques specific to this algorithm. Finally, we build an online decision tree learning system on FPGA with system-level optimizations. Our design remarkably reduces memory and computational demand, while achieving 0.05\% -- 12.3\% improvement in accuracy and 384$\times$ -- 1581$\times$ speedup in execution time over the state-of-the-art design.

\balance
\bibliographystyle{IEEEtran}
\bibliography{ref}
	
\end{document}